# The Dynamically Extended Mind

## A Minimal Modeling Case Study

Tom Froese, Carlos Gershenson, and David A. Rosenblueth
Departamento de Ciencias de la Computación
Instituto de Investigaciones en Matemáticas Aplicadas y en Sistemas (IIMAS)
Universidad Nacional Autónoma de México (UNAM)
Apdo. 20-726, 01000 Mexico D.F., Mexico
t.froese@gmail.com, cgg@unam.mx, drosenbl@unam.mx

*Abstract*—The extended mind hypothesis has stimulated much interest in cognitive science. However, its core claim, i.e. that the process of cognition can extend beyond the brain via the body and into the environment, has been heavily criticized. A prominent critique of this claim holds that when some part of the world is coupled to a cognitive system this does not necessarily entail that the part is also constitutive of that cognitive system. This critique is known as the "coupling-constitution fallacy". In this paper we respond to this reductionist challenge by using an evolutionary robotics approach to create a minimal model of two acoustically coupled agents. We demonstrate how the interaction process as a whole has properties that cannot be reduced to the contributions of the isolated agents. We also show that the neural dynamics of the coupled agents has formal properties that are inherently impossible for those neural networks in isolation. By keeping the complexity of the model to an absolute minimum, we are able to illustrate how the coupling-constitution fallacy is in fact based on an inadequate understanding of the constitutive role of nonlinear interactions in dynamical systems theory.

## I. Introduction

The Extended Mind Hypothesis (EMH), first formulated by Clark and Chalmers in 1998 [1], is continuing to receive much positive attention in the philosophy of cognitive science [2-5] and fits an embodied, enactive, and dynamical approach [6-9]. However, its core claim, that cognition can extend beyond the brain and into the environment, also continues to be criticized, especially for supposedly committing a "coupling-constitution fallacy" [10-12]. This criticism is based on the claim that when a part is coupled to a cognitive system this does not necessarily entail that this part is also constitutive of that cognitive system. With the growing incorporation of complexity theory into most areas of natural science [13], this kind of reductionist critique seems limited and outdated. And yet, even though the concept of systemic coupling is also foundational to dynamical systems theory, it is surprising that so far there have only been few attempts to evaluate this point of contention from such a more formal perspective. Nevertheless, reformulating the EMH in systemic terms is beginning to be considered as a promising response [8, 14, 15]. For ease of reference, we will refer to such a systemic perspective on behavior and cognition [16-18] as the Dynamical Systems Hypothesis (DSH).

As yet another alternative to orthodox functionalism in the philosophy of cognitive science, the DSH is losely affiliated with the EMH [19]. Nevertheless, it is still not clear just how compatible these alternative approaches are [20]. Specifically, while the EMH has largely developed out of orthodox functionalism, and therefore still retains many of its guiding assumptions, the DSH was originally derived from a different branch of cybernetics, one which is more compatible with ecological psychology and systems biology [21]. The essential difference between these two traditional frameworks becomes evident when we consider their divergent opinions on what constitutes a paradigmatic cognitive system, i.e. a computer running algorithms on data, versus an organism engaged in an ongoing interaction within its environment.

These two competing intuitions have resulted in different theories about the nature of cognitive extension. The EMH in effect accepts that cognition is something that primarily occurs in the brain, and secondarily asks about possible exceptions to this internalism, such as during tool-use. For the DSH, on the other hand, the mind was never located only inside the head in the first place [14]. Instead, cognition is primarily conceived of as a form of viable conduct by an agent in an environment [22]. Cognition is therefore treated as a relational process spanning the nonlinear interactions of brain, body, and environment [23], including other agents [24]. And although it helps to analyze the different contributions made by the individual components, the properties of the overall process cannot simply be reduced to one of the isolated components alone. In other words, the DSH takes the notion of an extended mind as its starting point, rather than as a curious exception.[1] A similar conclusion is also starting to be advocated in response to the challenge of the coupling-constitution fallacy for the EMH [26-28].

How does the DSH avoid the coupling-constitution fallacy? First of all, according to the DSH the very question of a spatial location of cognition is misguided. For example, consider the fact that the current paragraph is placed on the right-hand side in relation to the first column of text. But where is this relation of "being on the right-hand side" itself located? It is not on the right-hand side; since it is a relational property that compares

---

T.F. was supported by a postdoctoral fellowship of UNAM. C.G. was partially supported by SNI membership 47907 of CONACyT, Mexico. This work was partially funded by DGAPA PAPIIT grant IN113013.

[1] The DSH, like the enactive approach to cognitive science, is instead faced with the challenge to explain abstract forms of cognition, especially if they do not seem to involve interaction with the environment in any essential manner. We will not try to address this challenge here, but see e.g. [25].

two phenomena, it itself is not located anywhere. The same reasoning applies to cognition if we conceive of it as a kind of adaptive relationship between an agent and its environment. If cognition is a relational phenomenon, it logically cannot be located inside the brain (or anywhere). The same can be said about our conscious experience, which involves a perspectival relationship between body and environment. However, given that this DSH response depends on accepting precisely that which is at stake in the first place, i.e. an extended view of the mind, it is not much help against the charge of a coupling-constitution fallacy. Even if the critics accepted that cognition is a relational phenomenon, they would still want to hold on to the claim that the physical mechanisms underlying cognition (or conscious experience [29]) are limited to the activity of the nervous system. Thus, a better response to the critics would be to demonstrate that, even if we accept their internalist starting point, a proper understanding of neuronal activity will force us to accept an extended view of the mind nonetheless.

In the following we provide a proof of concept in the form of a minimal model of an embodied agent. We show that the artificial nervous system (ANS) of this agent is a qualitatively different kind of dynamical system when it is embodied and situated in an environment, compared to when it is considered in isolation. Moreover, in order to counter the potential charge that the agent's body and/or the environment are only serving as external input or secondary scaffolding for what essentially remains an isolated ANS, we demonstrate that the embodied and situated ANS has a formal property that is impossible to be generated by any isolated part of the model. If we take the class identity of a dynamical system to be defined by its organization [30], then this change of identity of the ANS from one type of system to another, is only explainable as an emergent outcome of nonlinear coupling between ANS, body and the environment subsystems. From the perspective of dynamical systems theory, this formal mathematical result is not surprising:

"[…] when the agent and the environment are nonlinearly coupled, they, together, constitute a nondecomposable system, and when that is the case, the coupling-constitution fallacy is not a fallacy. In other words, the coupling-constitution fallacy is only a fallacy when the coupling is linear." ([8], pp. 31-32)

In general, nonlinear coupling tends to transform the identity of the interacting components, and to assume otherwise would be instead to commit an error that we might call the "coupling-reduction fallacy". This reductionist fallacy was characteristic of the classical scientific worldview from Galileo and Newton onwards, but such linear thinking is rapidly being replaced by a worldview based on complexity [13]. In conclusion, by arguing that the organism-environment interactions are best described in terms of nonlinear coupling between dynamical systems, we reject the couping-constitution fallacy as unfounded.

## II. Theory

We took our initial inspiration from Walter's [31] famous robot "tortoises", which were some of the first mobile robots. In spite of only having a simple reactive control architecture, and only touch and light sensors, the robots were capable of exhibiting a variety of lifelike behaviors, especially in interaction with each other, that lent themselves to interpretation in cognitive terms.

Similar to later examples of situated robotics and models of minimal embodied cognition (e.g. Brooks [32] and Beer [33]; for a review, see [34]), such artificial agents are powerful tools for the EMH and DSH. The analysis of what is happening in terms of the whole systems is tractable, while still strongly making the point that cognition is primarily a relational and distributed process [35]. We follow this tradition in keeping our model of embodied agents as minimal as possible to better illustrate the transformative effects emerging from nonlinear coupling between dynamical systems.

In particular, our aim is to build a model of an embodied agent, whose artificial nervous system (ANS) has mathematical properties that are in principle impossible for it to have in isolation. The motivation for this criterion is the need to go beyond a demonstration of how an agent's situatedness within a sensorimotor loop modulates the internal activity of the ANS, but can transform the ANS into a qualitatively different kind of system altogether. Chaos is a suitable property for measuring such a transformation. It has been mathematically proven that it is impossible for a phase plane to exhibit chaotic dynamics (the Poincaré-Bendixon theorem); a smooth dynamical system must be at least three-dimensional, i.e. consisting of at least three partial differential equations, to be capable of exhibiting chaotic activity [36]. In other words, if an ANS with less than 3D is nonlinearly coupled with other non-chaotic systems, and its internal neural activity spontaneously becomes chaotic, then an explanation of this property as resulting from an extended process of interaction cannot be accused of committing the coupling-constitution fallacy. This minimal proof of concept is intended as an illustration. It does not entail that all neuronal activity is constituted by nonlinear coupling with body and environment in practice (some activity is mainly endogenous to the brain), but it clearly shows that there is nothing mysterious about cognitive extension in principle.

Related work has been done by Beer [37], who showed that an isolated 2D dynamical system of a particular kind can be transformed into a chaotic one by externally modulating it with a sinusoidal input having appropriate amplitude and frequency. But in this case it could still be argued that the input signal is only a form of external scaffolding that is independent of the system itself. It cannot be denied that this signal has a causal impact on the activity of the system, but this is not sufficient to count it as a constitutive part of that system. An overlenient criterion of inclusion raises the worry that anything that has an effect on the system must be included as being part of that system. But this is not so. Mathematically, the distinction can be expressed as the difference between an external parameter and an internal variable of the system. Nevertheless, a related worry of "cognitive bloat" has been charged against the EMH, so it is important to find a principled manner of avoiding this problem of over-inclusiveness.

We follow the enactive approach to cognitive science [7] in defining the systemic identity of a network of processes in terms of operational closure. More precisely, in order for a process to be included as a part of the system it must enable, and be enabled by, at least one other process of that system [38]. A paradigmatic example is the self-producing network of metabolic processes that physically bring forth the organism, i.e. autopoiesis [39]. Some proponents of the EMH have raised the worry that this concept of operational closure is tied to an internalist theory of mind [40], but this worry confuses the organism's spatial boundaries with its organizational limits [41]. There is nothing preventing the operational closure of the living from extending into the environment, as long as the condition of co-dependence applies. Regarding Beer's example of a 2D system with chaotic activity, we know that this activity depends upon properties of the input signal, but this signal does not in turn depend upon the activity of the system. Thus, the signal does not form a part of that system. In order to count as a genuine example of extension, the input signal itself must be dependent on the activity of the system as well.

We therefore set out to create a minimal agent-based model where this operational closure criterion of systemic inclusion is satisfied. Previous work in evolutionary robotics has shown that the interaction process between two embodied agents can transform the state space of their ANS, often implemented as a Continuous-Time Recurrent Neural Network (CTRNN), such that new behaviors emerge [42-44]. This finding fits well with evidence from experimental psychology and neuroscience, which also supports the possibility of a socially extended mind [45-48]. Accordingly, we hypothesized that an agent's ANS state space may become extended across both CTRNNs during some forms of interaction [24], such that even agents with a 2D or lower dimensional ANS can exhibit chaotic neural activity. The aim was for the agents to interact with each other so as to mutually complement their internal dynamics to overcome the restrictions that in principle apply to isolated 2D systems. However, as will be described in the next section, it turned out that our criterion can already be satisfied by a much simpler model of interacting agents.

### III. METHODS

In this section we briefly describe the methods used in order to create a minimal model of a chaotic embodied agent. We began work on our model as a replication of Di Paolo's [49] model of two acoustically coupled embodied agents, which was created using an evolutionary robotics method [50-52]. Details of this model and of our modifications are provided to ensure replicability of the approach. Some readers may wish to skip the technical details of how our model was created and proceed to the results presented in the next section.

*A. The body and environment*

The body of each agent is modeled as a circular object of radius $R = 4$ with two motors placed on parallel sides and two sound sensors symmetrically positioned at 45 degrees to the motors. This placement of the sensors introduces a back/front asymmetry. In contrast to Di Paolo's model, the agent makes a constant sound, whose source is located at the center of their body. Because the agent's own sound is constant, it is ignored by its sensors. The motors can rotate backwards and forwards, thereby moving the agent in a 2D unstructured and unlimited arena. For purposes of simplicity, the body is treated as a small rigid object with a very small mass, so that the motor output is the tangential velocity at the point of the body where the motor is located. Translational movement of the whole agent is calculated as the velocity of its center of mass (i.e. the vector average of the two motor velocities), and the rotational movement is calculated as the angular speed (i.e. the difference of the tangential velocities divided by the diameter of the body). The model does not include inertial resistance to either form of movement.

The agents are placed in an unlimited 2D environment that is empty apart from the agents' bodies. The agents move freely in this arena except when they collide. Collisions are modeled as point elastic, i.e. without energy loss and without effect on the angular velocity of the bodies. During a collision, an agent moves in a direction which is not the one specified by its two motors, but which corresponds to a displacement that conserves the momentum of the whole system. The bodies of both agents are taken as identical so that the result of an elastic collision is the instantaneous 'swapping' of the velocity vectors at the center of mass. However, due to the lack of inertia, the agents recover control of their movement immediately after the collision. The body circumference is taken as frictionless so that the angular velocities do not change during collisions. No touch sensors are used.

Sound is modeled as an instantaneous additive field of a single frequency with time-varying intensity, which decreases with the square of the distance from the source. For simplicity, the model excludes the effects of time-delays and differences between frequencies of sound production (e.g. Doppler effect, differential filtering). To help the agents in performing spatial discrimination with their sound sensors, these are placed at a distance from each other on the outside of the body. The two sensors will thereby be influenced by different intensities from the same external source, and this relative activity provides a basis for spatial discrimination. In addition, sound is treated as high-frequency such that the sound's penetration of the body is a source of attenuation. The amount of attenuation is related to the angular position and movement of the listening agent (the agent's own constant sound is ignored). The body's 'sound-shadowing' mechanism is implemented as a linear attenuation without refraction proportional to the distance traveled by the signal within the body, $D_{sh}$. This distance is given by

$$D_{sh} = D_{sen}(1 - A), \qquad 0 \leq A < 1$$

$$A = \frac{D^2 - R_0^2}{D_{sen}^2}$$

where $D_{sen}$ is the distance between the sound source and the sensor, and $D$ is the distance between the source and the center of the body. If $A \geq 1$, there is a direct line between source and sensor, and so $D_{sh} = 0$. For $A = 1$ the agent's sensor, the center of its body and the external source form a right triangle. The maximum value of $D_{sh}$ is given when the sensor is directly adjacent to the external source ($D_{sh} = 2R$). The intensity of the attenuated sound signal is calculated by first determining the intensity of the signal at the position of the sensor in the usual way, namely by applying the inverse square law without any attenuation, and then multiplying by an attenuating factor that goes linearly from 1 when $D_{sh} = 0$ to 0.1 when $D_{sh} = 2R$. The process is done once for each sensor at each time step.

*B. The brain*

We chose to follow the standard procedure in evolutionary robotic and therefore modeled the agent's ANS with a CTRNN [23]. A CTRNN has the following state equation:

$$\tau_i \dot{s}_i = -s_i + \sum_{j=1}^{N} w_{ji} \sigma\left(g_j(s_j + \theta_j)\right) + I_i$$

$$\sigma(x) = \frac{1}{1 + e^{-x}}$$

$$i = 1, \ldots, N$$

where *s* is the state of each artificial neuron in analogy with the cell potential, $\tau$ is its time decay constant, $w_{ji}$ is the strength of the connection from the $j^{th}$ to the $i^{th}$ neuron, *g* is a gain, $\theta$ is a bias term, $\sigma(x)$ is firing rate modeled by the standard logistic activation function, and *I* represents an external stimulus from a sensor modeled as a simple input current. The initial neural architecture consisted of three fully interconnected neurons (*N* = 3). We chose to exlude the gain parameter by fixing it at a constant 1. We began the artificial selection process with such a 3D system rather than a 2D system because this redundancy can facilitate the evolution of interesting behavior. The extra neuron served the role of interneuron and was only connected with other neurons. The other two neurons receive inputs from both of the agent's sensors, where each input connection is multiplied by an input gain parameter. These same two neurons are also connected to the motors in a one-to-one manner. More precisely, neuron *i*'s output $\sigma_i$ is transformed into the motor velocity $v_i$ as follows:

$$v_i = o_i * ((\sigma_i * 2) - 1)$$

where the neuron's original output range ($0 \leq \sigma_i \leq 1$) is first translated to the range [-1, 1] and then mulitplied by an output gain parameter $o_i$. This translation allows the same neuron to produce both forward rotation and backward rotation. Before calculating an agent's change in position and orientation, the velocities were multiplied by the size of the integration time step. Numerical integration of each CTRNN was implemented using the fourth order Runge-Kutta method with an integration step of 0.1.

*C. Evolutionary algorithm*

A form of rank-based selection genetic algorithm was used as an optimization procedure of the agent's behavior. The size of the population was fixed to 100 solutions evolving for up to around 1000 generations, although the process was generally interrupted beforehand because fit behaviors typically evolved after a few hundred generations. A solution was implemented as a real-valued vector of fixed dimension, which encoded all CTRNN parameters (i.e. weights, gains, and biases) of one agent. Agents were structurally identical. At the beginning of a search, the population of vectors was initialized to random values in the range [-1, 1]. Search was constrained such that the vector values are clipped to this range. Each component of the vector was linearly scaled to the appropriate interval of the CTRNN parameter. The connection strengths were chosen from the interval [-8, 8], biases from [-3, 3], input gains from [-10, 10], output gains from [-2, 2], and time decay constants from [1, 50].

Selection pressure was set so that the best solution produces 1.3 offspring on average per generation. No elitism was used; all solutions are replaced after each generation. New offspring are created by applying a mutation operator to the parents. The mutation operator is Gaussian in nature, perturbing the parent solution with a mutation vector (i.e. a vector of real numbers specifying the changes for each component of a parent vector). The mutation vector's direction is uniformly distributed on the unit hypersphere and its size is normally distributed with a mean of 0 and variance of 0.2. No crossover operator was used.

A solution was evaluated for 10 trials, each with different starting positions and orientations (see below). The total fitness of a solution was calculated as an inverse weighted sum of the 10 scores. By assigning more weight to less optimal scores, the evolutionary process is forced to generalize more; otherwise it is likely to focus on optimizing only a subset of easy conditions at the expense of more difficult conditions. At the beginning of each trial the orientations of the agents were randomly selected from a uniform distribution with range [0, 2π]. The initial (*x*, *y*) position of one agent was randomly selected from a uniform distribution with ranges ([-20, -5], [-20, 20]), and the other's from ranges ([5, 20], [-20, 20]). This randomization of initial conditions ensured that the evolving behavioral strategy could not simply rely on consistent starting parameters, but had to be interactive and flexible. The initial ranges of the two agent's positions were chosen so as to avoid bodily overlap. The neural activations of each agent were initialized to 0.

Each trial lasted for 300 units of time, and a solution was rewarded in relation to how well the agents managed to spend their time close to each other. At each time step the difference between the initial distance from one agent's center of body to the other's center and their current distance was cummulatively added. This relative measure of distance was used to avoid biasing the fitness score according to the initial starting distance. The trial's final fitness score was made dependent on the distance at each time step during the trial, rather than only the final distance, so that the agents are forced to head toward each other as quickly and consistently as possible. This simple evaluation function was chosen to evolve the agents to engage in an interaction process like Walter's [31] robotic tortoises.

The agents were readily evolved to locate each other from any starting configuration by relying on the signals provided by their sound sensors. After a reliable strategy was found, the 3[rd] interneuron was pruned from the CTRNN by setting all of its incoming and outgoing connection strengths to 0. The resulting 2D system was then used as a seed for another population of solutions and further optimized by artificial evolution for a few dozen generations until the original strategy had been more or less reestablished. Because we wanted to better understand the minimal neural basis of this behavior, we decided to also prune the 2[nd] neuron in the same manner. That neuron's output to the 2[nd] motor was thereby fixed to a constant σ(θ), i.e. the logistic activation function applied to the neuron's bias term. The new 1D system was again further optimized by using it as a seed to create another population so as to continue evolution for a few dozen generations until fitness was sufficiently regained. To our surprise we could find suitable solutions even though the agents were reduced to one artificial neuron each.

## IV. RESULTS

These embodied agents, each with only a 1D artificial nervous system (ANS), were the simplest possible nonreactive agents for this task[2]. As should be expected, since even reactive agents can succeed in a variety of homing tasks [54], they are still capable of finding each other. The 1D agents accomplish this task by spiraling in the direction of the strongest signal, and then colliding together (see Fig. 1 for an illustrative example).

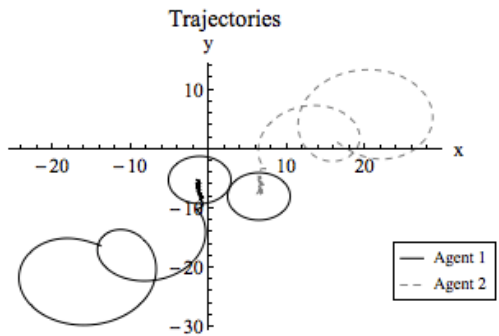

Fig. 1. Illustration of the trajectories taken by two acoustically coupled agents as they move in a spiral pattern and approach each other. The CTRNN of each agent consists of only 1 neuron (in isolation it is a 1D continuous system).

We then analyzed the properties of the time series of neural activity of 'agent 1' that was recorded during the trial shown in Fig. 1. What was immediately evident from visually examining the spiraling trajectories of the agents' interaction process was that their changes in neural activation were converging on a relatively regular oscillatory pattern. We confirmed this pattern by plotting the agents' neural activity (see Fig. 2).

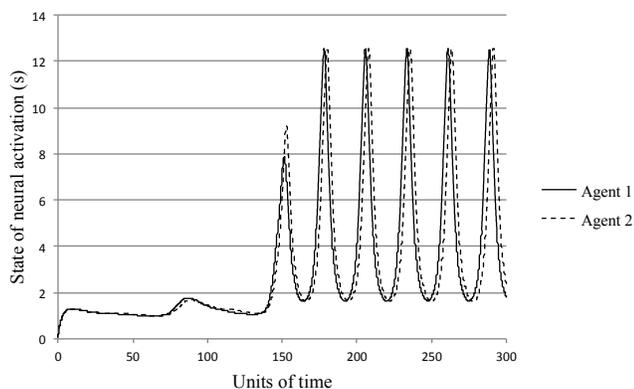

Fig. 2. Illustration of agents' change of neuronal activation ($y$-axis) over time ($x$-axis). Although the agents' CTRNNs are 1D in isolation, nonlinear coupling with each other via their embodiment enables a complex pattern of activity: an oscillation that never fully converges to a fixed limit cycle.

This oscillatory pattern is already an interesting result for our purposes because limit cycles are not possible for a 1D continuous system that is not nonlinearly coupled to another system [37]. In principle, isolated 1D continuous systems will always converge on a fixed-point attractor. More importantly, in the case of our model it cannot be argued that the oscillatory signal is externally provided for the agents, because neither one of the two agents is capable of producing such a signal alone, nor is it given by the external environment. The oscillation is an emergent product of the interaction between the two agents and their environment. The observation that the oscillation did not seem to be repeating precisely suggested that we may have found chaotic neural activity. However, in order to confirm this intuition we needed a more formal analysis of the pattern.

Following standard practice for determining the underlying systemic properties of a nonlinear time series [55], we had to calculate the CTRNN's effective dimensionality as well as its largest Lyapunov exponent. The latter is used for evaluating whether a time series is chaotic, with positive exponent values being a clear sign of chaos. In comparison, Beer ([37], p. 495) observed the smallest isolated CTRNN with chaotic dynamics to be a 3D system. The largest Lyapunov exponent was 0.010 and the effective Lyapunov dimension was 2.021, meaning that it was mildly chaotic. In terms of coupled nonlinear systems, Beer ([37], p. 502) found that driving a 2D system with an external periodic signal can produce chaotic dynamics. In his example the largest Lyapunov exponent was 0.023 and the Lyapunov dimension was 1.157.

The CTRNN in our model differs from Beer's examples in that it is even smaller (1D instead of 2D), and no independent input signal is provided. In order to obtain a better view of the long-term properties of the system, we ran the model for 3000 units of time instead of the 300 units used during the evolutionary optimization. Using a measure of the false nearest neighbor to calculate the effective embedding dimension of one agent's internal neural activity, we found the underlying system to be 3D. According to the Poincaré-Bendixon theorem (see [56], p. 210), a continuous system has to be at least 3D in order for chaotic activity to be possible.

No random number generator was used during the run of a simulation, but it is always good to check for any unintended stochastic variation in the model. Using the determinism test introduced by Kaplan and Glass [57] we found the time series of activation to be highly deterministic (determinism factor $k = 0.93$); an absolutely deterministic time series would have been $k = 1$. The slight difference is probably due to the model's discrete step-by-step approximation of a continuous system). Following Strogatz's ([56], p. 323) tripartite definition of chaos as "*aperiodic long-term activity* in a *deterministic system* that exhibits *sensitive dependence on initial conditions*", the only thing left to do is to determine the neural activity's sensitivity to initial conditions. This can be done by calculating the neural time series' largest Lyapunov exponent, which needs to be positive. And indeed the largest Lyapunov exponent was found to be 0.091, which indicates that the system's dynamics are slightly more chaotic than Beer's examples.

These two analytic results are complements of each other, because smooth dynamical systems need to be at least 3D to exhibit chaotic dynamics. It also follows that the agent's behavior, of which the neurons are a component, must be extended through the environment, since each isolated CTRNN is maximally a 1D system. By way of comparison, the largest Lyapunov exponent of the neural activity of an isolated CTRNN is -0.414, i.e. neither chaotic nor even very complex. The qualitative difference in dynamics of the isolated CTRNN can also be intuitively seen by comparing Fig. 2 with Fig. 3. In the latter the neural activity quickly converges on a single fixed-point attractor and remains there for the rest of the trial.

---

[2] Strictly speaking, this is not entirely accurate. Even the behavior of an agent with a 0D ANS, i.e. with a reactive controller, is not reactive – as long as the agent is embodied [53]. We will return to the role of embodiment later.

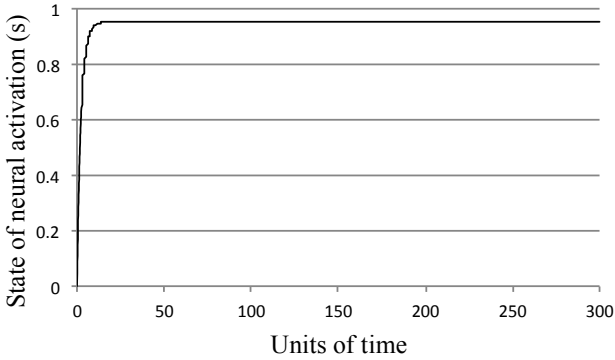

Fig. 3. Illustration of the time series of neural activation of the agent's isolated CTRNN. The 1D system quickly converges to a fixed-point attractor.

Thus, to our surprise, we managed to find a proof of concept of the validity of the EMH, based on the principles of the DSH, that was even more minimal than we had originally expected to be the case. Moreover, a closer investigation of the necessary elements of the chaotic neural activity gives support to a related hypothesis, namely that the mind is embodied. If we turn the agents into disembodied ghosts by removing the possibility of collisions, their complex pattern of activity can no longer be sustained and the dynamics converge on a fixed-point (Fig. 4). In other words, the agents' embodied orientation and position provide an additional source of dimensionality, which during nonlinear interactions can lead to chaos.

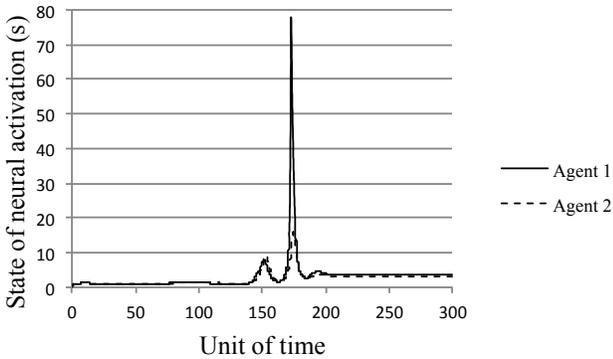

Fig. 4. Illustration of agents' change in neuronal activation ($y$-axis) over time ($x$-axis). The experimental conditions are the same as in Fig. 2, but without modeling the collisions between agents. Initially the two time series are the same as those of Fig. 2 until the time of the first bodily collision after about 175 units of time. In the absence of collisions the agents continue to move closer to each other such that agent 1's light sensor passes over the light-emitting center of the other agent (large neural activation spike for agent 1), until they come to rest such that their bodily centers are precisely on top of each other (both neural activations converge on a fixed-point attractor).

## V. DISCUSSION

Our model of agent-environment interaction was primarily intended to provide a word of caution against overestimating the actual force of the coupling-constitution fallacy, which has been the focus of much debate regarding the EMH. We have shown how the agent's behavior in the environment emerges as a result of the combined contributions of the various aspects of the overall model, i.e. two brain-body-environment systems in interaction with each other. We chose to model and analyze a minimal case of social interaction as an illustrative example of the dynamically extended mind, but the same arguments apply to embodied agents situated in a complex environment more generally. From the perspective of the DSH, which proposes a distributed view of cognition as the default mode of cognition, there is no coupling-constitution fallacy because properties of the sensorimotor interaction process cannot be reduced to that of the isolated components. We tried to illustrate this important insight with a minimal agent-based model, which gave us the following proof of concept:

1. A continuous-time recurrent neural network (CTRNN) is a type of continuous dynamical system.

2. The isolated (and therefore non-extended) CTRNN, that was intended for nonlinear coupling in our model, is 1D only.

3. The activity of an isolated continuous dynamical system can only be cyclical if its phase space is at least 2D, and its activity can only be chaotic if its phase space is at least 3D (Poincaré–Bendixson theorem).

4. The non-isolated CTRNN realized in our model exhibits oscillatory dynamics (i.e. approach toward limit cycle) as well as chaotic dynamics (i.e. positive largest Lyapunov exponent).

5. It logically follows from the above that the phase space of the agent's CTRNN must be explained in terms of the whole brain-body-environment-body-brain system.

But how does this relate to the EMH, which begins with the assumption that cognition first of all takes place in the brain? If all that we were able to show with our model is that the agent's ANS is constitutive of the emergent properties of an extended interaction process, this would not necessarily entail that the activity of the ANS itself was also constituted by, rather than merely caused by, this interaction process. It is a good strategy for the EMH to highlight that ongoing interaction between a cognitive agent and environment results in a novel, mutually encompassing process with new properties of its own [15, 26]. This appeal to emergence is an important part of the response to the reductionist critics. But to demonstrate that some aspects of the interaction process can be constitutive of the internal activity of the agent, we need to go further. We had to show that the time series of neural activity had new properties, e.g. Lyapunov dimensionality and the largest Lyapunov exponent, that were impossible to obtain for that CTRNN in isolation. Moreover, we required that this form of extension was not only due to an independent cause, but was part of the activity of the CTRNN. We therefore continue our proof of concept:

6. The non-isolated CTRNN's output is determined by its input, albeit mediated by its internal activity, while this input is determined by its motor output, albeit mediated by bodily and environmental (including social) activity.

7. It logically follows from the above that the non-isolated CTRNN's additional neural complexity is partially constituted by its own sensorimotor and social coupling.

Nevertheless, it could be argued that this proof of concept is of limited value when considering human cognition in the real world. For one thing, it is based on an extremely minimal model. Furthermore, we did not systematically search the space of solutions, and so we cannot know how representative this solution is. Certainly, we do not expect all nonlinearly coupled 1D continuous dynamical systems to exhibit a similarly high dimensionality and chaos. And yet the general message of the model does have some empirical support. For example, when two people oscillate their limbs (fingers or legs) while in visual contact, they start to spontaneously synchronize their motions as the frequency of oscillations increases [9, 58]. In fact, such synchronization is a widespread phenomenon, including during vocalization [59]. Moreover, acoustic social interactions can evoke profound changes in a person's brain and body: guitar duets are preceded and accompanied by inter-brain oscillatory couplings [60], and choir singing is related to inter-personal cardiac and respiratory phase synchronization [61].

Nevertheless, we do not expect that the model captures all that is essential about human social interaction in particular; for example, it applies equally well to single-cell interactions, bird duets, and robotic multi-agent systems [49]. But this generality of the model's results is precisely what is at stake, and it is why we chose to analyze a minimal dynamical system where the underlying principles are not obscured by the complexity of the phenomenon. The complexity of a human brain makes it hard to know how it is interactively transformed and extended in practice, but at least there is no such difficulty in principle. The claim that the dynamical organization of a system is changed during a nonlinear interaction with other systems is formally based on dynamical systems theory, in particular on the basic distinction between so-called "autonomous" (isolated) and "non-automous" (coupled) systems[3]. Add to this mathematical insight the fact that sensory input and motor output form a closed sensorimotor loop, and the extended mind hypothesis takes on the form of a universal principle of life and mind.

Nevertheless, there is something right about the traditional folk-psychological intuition that our mind is individuated in our body, at least that is how it seems to us from our first-person perspective most of the time. This is a phenomenon that clearly deserves explanation. The introspective observation that we experience ourselves as having a relatively stable personal identity, an identity that is not always continually transformed and dissolved during our interactions in the world (unlike for some people with schizophrenia), is something that must still be accounted for from the perspective of the DSH. The real challenge facing the DSH, therefore, is not to explain how cognition sometimes becomes extended, but rather how our relatively stable sense of self is maintained even in the face of continous sensorimotor and social extension. Accordingly, the nervous system's endogeneous activity deserves a closer look, in particular in terms of those processes that allow a temporary decoupling from ongoing interactions [63].

## VI. Conclusion

As a variant of Cartesian dualism, the computational theory of mind has traditionally assumed that cognition is something that only takes place within the head. However, during the last two decades this internalist boundary has been questioned by embodied, enactive, and extended theories of mind. In response critics have argued that the extended mind hypothesis is based on a coupling-constitution fallacy, i.e. that when a cognitive process in the head is coupled with a part of the environment, that part is neither necessarily nor typically constitutive of the cognitive process. However, this philosophical critique is not supported by nonlinear dynamical systems theory..

We have used an evolutionary robotics approach to create a minimal model that illustrates some of the transforming effects of nonlinear agent-environment coupling both on an embodied agent's external behavior and on its internal artificial nervous system. A dynamical systems approach thus has the potential to formally save the extended mind hypothesis from its critics. Nevertheless, this mathematical response comes at a price for mainstream cognitive science: it predicts interactive cognitive extension to be the primary and default condition of mind (in evolutionary, developmental, and behavioral terms), rather than an occasional exception to an otherwise internal mind.

ACKNOWLEDGMENT

The computer model was implemented with the help of the Evolutionary Agents C++ software package v1.1.3, which was made available by Randall D. Beer. The Lyapunov analysis of the time series data was performed using the Nonlinear Time Series Analysis programs provided by Matjaž Perc. (Support by funding agencies is acknowledged separately on the bottom of the first page of this article).

---

[3] This mathematical use of the term "autonomy" should not be confused with uses of the term found in biology and other disciplines [62].